\pdfoutput=1
\documentclass[10pt, titlepage]{article}
\usepackage{amsmath,amsthm,amssymb, listings, tikz, pgf, mathtools}
\usepackage[top=1.25in, bottom=1.25in, left=1.25in, right=1.25in]{geometry}
\usepackage{pgf}
\usepackage{tikz}
\usepackage{subcaption}

\usepackage{multicol}
\usepackage{enumitem}
\usepackage{sectsty}
\subsectionfont{\normalsize}
\subsubsectionfont{\small}
\usepackage{listings}
\usepackage{changepage}

\usepackage{float}

\usepackage{footnote}
\makesavenoteenv{tabular}
\makesavenoteenv{table}

\usepackage[linesnumbered,ruled]{algorithm2e}

\usepackage{csquotes}
\usepackage{gensymb}
\usepackage[font=footnotesize]{caption}

\theoremstyle{definition}
\newtheorem{definition}{Definition}[section]

\usetikzlibrary{arrows,automata}
\usepackage[latin1]{inputenc}

\newcommand\mb{\mathbf}

\DeclareMathOperator*{\argmax}{argmax}

\DeclarePairedDelimiter\floor{\lfloor}{\rfloor}

\usepackage{listings}
\usepackage[hidelinks]{hyperref}

\author{Kaiyu Zheng}
\date{April 2017}

\begin{document}

\begin{titlepage}
\centering
\topskip0pt
\vspace*{\fill}
  \huge Learning Large-Scale Topological Maps Using Sum-Product Networks\normalsize

  \vspace{0.5in}

  \Large Kaiyu Zheng\large

  \vspace{1.2in}

  A thesis submitted in partial fulfillment of the requirements for the degree of

  \vspace{0.1in}

  Bachelor of Science with Departmental Honors

  \vspace{0.8in}
  University of Washington

  2017

  \vspace{0.4in}
  Advisors:

  Dr.~Andrzej Pronobis

  Prof.~Rajesh Rao

  \vspace{0.8in}
  Program Authorized to Offer Degree:

  \vspace{0.1in}

  Paul G. Allen School of Computer Science \& Engineering\normalsize
\vspace*{\fill}

\end{titlepage}

%%%%%%%%%%%%%%%%%%%%%%%%%%%%%%
% For ...
%%%%%%%%%%%%%%%%%%%%%%%%%%%%%%
\newpage

%%%%%%%%%%%%%%%%%%%%%%%%%%%%%%
% Abstract
%%%%%%%%%%%%%%%%%%%%%%%%%%%%%%
\newpage
\pagenumbering{Roman} % Start roman numbering
\begin{center}\Large \textbf{Abstract}\normalsize\end{center}
In order to perform complex actions in human environments, an autonomous robot needs the ability to understand the environment, that is, to gather and maintain spatial knowledge. Topological map is commonly used for representing large scale, global maps such as floor plans. Although much work has been done in topological map extraction, we have found little previous work on the problem of learning the topological map using a probabilistic model. Learning a topological map means learning the structure of the large-scale space and dependency between places, for example, how the evidence of a group of places influence the attributes of other places. This is an important step towards planning complex actions in the environment. In this thesis, we consider the problem of using probabilistic deep learning model to learn the topological map, which is essentially a sparse undirected graph where nodes represent places annotated with their semantic attributes (e.g. place category). We propose to use a novel probabilistic deep model, Sum-Product Networks (SPNs) \cite{poon2011sum}, due to their unique properties. We present two methods for learning topological maps using SPNs: the place grid method and the template-based method. We contribute an algorithm that builds SPNs for graphs using template models. Our experiments evaluate the ability of our models to enable robots to infer semantic attributes and detect maps with novel semantic attribute arrangements. Our results demonstrate their understanding of the topological map structure and spatial relations between places. 
%%%%%%%%%%%%%%%%%%%%%%%%%%%%%%
% List of Figures
%%%%%%%%%%%%%%%%%%%%%%%%%%%%%%
%% \newpage
%% \listoffigures

%%%%%%%%%%%%%%%%%%%%%%%%%%%%%%
% List of Tables
%%%%%%%%%%%%%%%%%%%%%%%%%%%%%%
%% \newpage
%% \listoftables

%%%%%%%%%%%%%%%%%%%%%%%%%%%%%%
% Table of Contents
%%%%%%%%%%%%%%%%%%%%%%%%%%%%%%
\newpage
\tableofcontents
 
%%%%%%%%%%%%%%%%%%%%%%%%%%%%%%
% Introduction
%%%%%%%%%%%%%%%%%%%%%%%%%%%%%%
\newpage
\pagenumbering{arabic} % Switch to normal numbers
\section{Introduction}
The fundamental value of robotics is two-fold. First, robots can take risks in the place of humans, do what humans cannot or struggle to do. Second, robots can assist humans to achieve their goals in a more efficient and effective way. The last several decades saw an explosion of interest and investment in robotics both in commercial applications and academic research \cite{pagliarini2017future}, and the field of autonomous robots is expected to be a topic of continuous heavy research.  In this thesis, we concentrate on mobile robots in indoor environments. Mobile robots are desirable to fulfill both aspects of the value, especially the latter; They have the potential to provide various kinds of services, and recently attempts have been made to apply mobile robots to real-world problems such as helping older people \cite{jayawardena2010deployment}, guiding passengers in airports \cite{triebel2016spencer}, and telepresence \cite{matsuda2016scalablebody}. However, these robots are still far from fully autonomous, and only perform limited actions with specially designed tasks.

To enable autonomous mobile robots to exhibit complex behaviors in indoor environments, it is crucial for them to form an understanding of the environment, that is, to gather and maintain spatial knowledge. The framework that organizes this understanding is a spatial knowledge representation. Additionally, due to the uncertainty in human environments, it is also crucial for the robots to be able to learn about the spatial knowledge representation in a probabilistic manner. In this thesis, we use the Deep Affordance Spatial Hierarchy (DASH) \cite{pronobis2017deep}, a recently proposed hierarchical spatial representation that spans from low-level sensory input to high-level human semantics.  The ultimate goal of our research is to use a unified deep generative model to capture all layers of DASH, an approach fundamentally different from the traditional where an assembly of independent spatial models exchange information in a limited way \cite{pronobis2016learning}. 
Specifically, in this thesis, we focus on learning the layer of topological map, a mainstream graphical representation of the information of places and their connectivity in a full floor map. %\cite{remolina2004towards}.
We have chosen to use Sum-Product Networks (SPN) \cite{poon2011sum}, a new class of deep probabilistic models. It is particularly well-suited for our purpose primarily because of three reasons. First, it is a deep generative model, therefore probabilistic by nature. Second, it guarantees tractable training and inference time with respect to the size of the network, a property important in robotics where real-time inference and planning is often required. Third, it is simple to combine multiple SPNs into a single unified SPN with interpretable layers, which naturally matches our purpose of learning a hierarchical representation.

Learning a topological map means learning the structure of the large-scale space and dependency between places, for example, how the evidence of a group of places influence the attributes of other places. This is an important step towards planning complex actions in the environment. By itself, it is also useful for tasks such as inferring missing information of a place based on the surrounding places. In the context of DASH, it is can be used for correcting the classification results of the local environment model\footnote{In DASH, above the layer of low-level sensory input, there is a layer called ``peripersonal layer'' which converts sensory input into robot-centric representation of the immediate surroundings, i.e. local environment. More details about DASH is given in section \ref{section:dash}.}. Although there has been considerable effort in the extraction of topological maps \cite{friedman2007voronoi}\cite{ranganathan2011online}\cite{shi2010online}\cite{tomatis2003hybrid}, only few works is tried to learn it with semantic information in the past two decades \cite{aydemir2012can}\cite{friedman2007voronoi}\cite{mozos2006supervised}.

%\subsection{Challenges}

There are three main challenges in this work. First, topological maps have varying sizes (number of places and connections), because the underlying environment may have different dimensions. Second, it is not clear how graphs such as topological maps should be modeled\footnote{In this thesis, ``modeling'' a topological map means the same as ``learning'' it.} using an SPN. Third, it is difficult to obtain topological maps for training, therefore the model needs to learn from small datasets.

%\subsection{Contributions}

We address the above challenges through our contributions. We present two methods to use SPNs to model topological maps: the place grid method, and the template-based method. Both methods aim to deal with the variability of topological maps but with different emphasis on the dependency between places. In the place grid method, we project the topological map onto a grid of fixed size, and generate an SPN structure on top of the grid. In the template-based method, we propose an algorithm to build SPNs for graphs based on template models. Each template is a basic graph structure that occurs throughout in the topological map. We train an SPN on a template, or a hierarchy of templates. For each topological map instance, we construct an instance-specific full SPN using the template SPNs as children, and do inference on the full SPN. To our knowledge, these are the first methods to use SPN to model a graph. We evaluate the two methods by experiments and our results show their capability of enabling robots to do inference of environment attributes with promising output, showing the understanding of the spatial relations between places. In addition, we created a large dataset with fully annotated localization, sensory information and virtual scans, a low-level geometry representation of robot-centric environment\footnote{We plan to publish this dataset to IJRR soon.}. This dataset is used to evaluate our methods.

\subsection{Constraints}
In this thesis, we focus on place category as the only semantic attribute for a place. Our rationale is the following. In terms of the learned model's potential usage, it would certainly help if other information such as types of objects in a place, or the low-level sensory information, is provided. However, our goal in this work is to evaluate the problem of learning topological maps in separation. The proposed methods naturally extend to the case where more types of semantic attributes or sensory information are considered. This is a direction for future work.

\subsection{Thesis outline}

\paragraph{Section \ref{section:related} - Related Works} We review related works in the use of topological maps as a part of spatial knowledge, learning of topological maps, and recent machine learning techniques used to model graphs.

\paragraph{Section \ref{section:background} - Background} We discuss details about the theoretical background related to our work. First, we describe details of Sum-Product Networks. Then, we present an overview of the Deep Affordance Spatial Hierarchy. Finally, we formalize the notion of topological map and describe the method we use to extract topological maps.

\paragraph{Section \ref{section:sol} - Problem Statement \& Proposed Solutions} We formalize the problem that we try to solve in this thesis. Then, we describe the details of the two presented methods that model topological maps using SPNs: the place grid method and the template-based method.

\paragraph{Section \ref{section:experiment} - Experiments} We describe the dataset, methodology, and results of our experiments.

\paragraph{Section \ref{section:conclusion} - Conclusion \& Future Work} We conclude our thesis by providing a summary of the presented methods and experiment results. Finally, we discuss improvements to be made in the future.

%%%%%%%%%%%%%%%%%%%%%%%%%%%%%%
% Related Works
%%%%%%%%%%%%%%%%%%%%%%%%%%%%%%
\newpage
\section{Related Works}\label{section:related}

The use of topological maps in navigation and mapping dates back to the work by Kulpers and Byun \cite{kuipers1991robot} in 1991, where they first considered the use of qualitative maps to avoid error-prone geometrically precise maps. Topological maps then began to be considered as a form of spatial knowledge \cite{kuipers2000spatial}. More recently, Beeson et.~al.~\cite{beeson2010factoring} proposed a hybrid approach of spatial knowledge representation that uses metrical representation for local environment and topological representation for large environment. Pronobis et.~al.~\cite{pronobis2010representing} outlined criteria for good spatial knowledge representation where the use of topological map to represent global environment is also promoted.

Aydemir et.~al.~\cite{aydemir2012can} investigated the statistical properties of indoor environments by analyzing two large floor plan datasets consisting of over 197 buildings and 38,000 rooms in total. They pointed out that indoor environments are not yet well understood. They discovered that local complexity remains unchanged for growing global complexity in real-world indoor environments. This finding provides important statistical support for our template-based method discussed later. They also proposed two methods for predicting room categories for a given topological map, but their methods rely on the maintenance of a large dataset of possible graphs, and do not involve learning of the actual topological map. In contrast, our methods can learn the dependency of place categories from a limited amount of data.

Sum-Product Networks (SPNs), proposed by Poon and Domingos \cite{poon2011sum} are a new class of probabilistic model that guarantee tractable inference. Despite their competitive or state-of-the art results on tasks such as image completion \cite{poon2011sum}, image classification \cite{gens2012discriminative}, and language modeling \cite{cheng2014language}, the advantages of SPNs have not been exploited much in robotics, besides \cite{pronobis2016learning}. Moreover, for neural networks, the problem of learning from data naturally structured as graphs has been on the radar for around a decade. Scarselli et.~al.~\cite{scarselli2009graph} proposed graphical neural networks, and recently there is an increase of intererst for generalizing convolutional neural networks (CNNs) beyond grid data such as images to model data in graph domains \cite{defferrard2016convolutional}\cite{kipf2016semi}, showing promising progress. In contrast, there has been no research on using Sum-Product Networks to learn similar data. This work offers a straightforward first step in broadening the scope of applications for SPNs.

Among the numerous attempts in extracting topological maps, some tried to learn the place categories in topological maps as well. Mozos and Burgard~\cite{mozos2006supervised} presented an approach which first classifies each point in the metric map into semantic categories using AdaBoost, then segments the labeled metric map into regions in order to extract a topological map. This approach is good for classifying nodes on the topological maps accurately, but the model does not learn the spatial relations between the categories since it does not support probabilistic inference. Friedman et.~al.~\cite{friedman2007voronoi} proposed an approach based on Voronoi random fields (VRFs), which also attempts to estimate the label of each node. A VRF is a conditional random field (CRF) converted from a Voronoi graph created from a metric grid map. Although this approach uses CRF, a discriminative probabilistic graphical model, it relies on pseudolikelihood and loopy belief propagation \cite{murphy1999loopy} to achieve approximate MAP inference, which may not converge to correct probability distribution. Besides, this approach relies on AdaBoost to learn features for place classification, similar to \cite{mozos2006supervised}. In this thesis, we use SPNs which guarantee efficient inference and correctness of the modeled probability distribution. Furthermore, as descrbed in section \ref{section:spn}, theoretical properties of SPNs support combining several SPNs into a unified SPN. This avoids the complexity of using multiple kinds of machine learning models trained for different purposes.

%%%%%%%%%%%%%%%%%%%%%%%%%%%%%%
% Background
%%%%%%%%%%%%%%%%%%%%%%%%%%%%%%
\newpage
\section{Background}\label{section:background}

In this section, I provide the background information for this thesis in detail. First, I describe in detail the properties of Sum-Product Networks (SPNs), how inferences are performed, and how the parameters and structure of the network can be learned. Next, I describe the Deep Affordance Spatial Hierarchy (DASH), the spatial knowledge representation that we use to enable environment understanding for mobile robots. Finally, I describe indoor topological maps in general, and how topological maps are generated in DASH.

\subsection{Sum-Product Networks}\label{section:spn}

\subsubsection{Definitions and Properties}

SPN, proposed by Poon and Domingos \cite{poon2011sum}, is a new class of probabilistic graphical models with built-in properties that allow \textit{tractable} inference, a major advantage over traditional graphical models such as Bayesian networks. The idea is built upon Darwiche's work on network polynomial and differentials in arithmetic circuit representation of the polynomial \cite{darwiche2003differential}. Here, we provide the definition of SPN and several of its key properties.

\begin{definition}[SPN]
\cite{poon2011sum}
Let $\mathcal{X}=\{X_1,\cdots X_n\}$ be a set of variables. A Sum-Product Network (SPN) defined over $\mathcal{X}$ is a rooted directed acyclic graph. The leaves are indicators $[X_p= \cdot ]$. The internal nodes are sum nodes and product nodes. Each edge $(i,j)$ from sum node $i$ has a non-negative weight $w_{ij}$. The value of a sum node is $\sum_{j\in Ch(i)}w_{ij}v_j$, where $Ch(i)$ is the children of $i$. The value of a product node is the product of the values of its children. The value of an SPN is the value of its root.
\end{definition}

We use $S$ to denote an SPN as a function of the indicator variables (i.e. the leaves). Let $\mb{x}$ be an instantiation of the indicator variables, a full state. Let $\mb{e}$ be an evidence (partial instantiation). For a given node $i$, we use $S_i$ to denote the sub-SPN rooted at $i$. Also, we use $x_{p}^a$ to mean $[X_p=a]$ is true, and use $\bar{x}_p^a$ to mean the negation, for simplicity. We define the following properties of SPN.

\begin{definition}[Validity]
An SPN is \textit{valid} if and only if it always correctly computes the probability of evidence: $S(\mb{e})=\Phi_S(\mb{e})$, where $\Phi_S(\mb{e})$ is the unnormalized probability of $\mb{e}$.
\end{definition}

\begin{definition}[Consistency]
An SPN is \textit{consistent} if and only if for every product node $i$, there is no variable $X_p$ that has indicator $x_p^a$ as one leaf of the sub-SPN $S_i$ and indicator $x_p^b$ with $b\neq a$ as another leaf.
\end{definition}

\begin{definition}[Completeness]
An SPN is \textit{complete} if and only if all children of the same sum node have the same scope. (The scope of an SPN is the set of variables in $\mathcal{X}$ that the indicators of an SPN are defined on.)
\end{definition}

\begin{definition}[Decomposability]
An SPN is \textit{decomposable} if and only if the children of every product node have disjoint scopes.
\end{definition}

The above definitions can constrain an SPN so that it is no longer an arbitrary multi-linear map from the indicators to a real number. Poon and Domingos proved that if an SPN is complete and consistent, then it is valid \cite{poon2011sum}. A more restricted theorem for validity is that if an SPN is complete and decomposible, then it is valid \cite{peharz2015theoretical}. We will apply the latter theorem in to solve our problem, since it is easier to guarantee that the children of a product node have disjoint scopes.

\paragraph{Hidden Variables}
Given a complete and consistent SPN $S$ and an arbitrary sum node $i$, we know:
\begin{equation}
S_i(\mb{x})=\sum_{j\in Ch(i)}w_{ij}S_j(\mb{x})
\end{equation}
If $\sum_{j\in Ch(i)}w_{ij}=1$, we can view the value of the sum node $i$ as summing out a hidden variable $Y_i$, where $P(Y_i=j)=w_{ij}$. Also, in this case, the partition function $Z_S=S(1\cdots1)=\sum_{\mb{x}}S(\mb{x})$=1, because all of the indicators have value 1, and the product nodes and sum nodes all output 1. It follows that the value of every sub-SPN is a normalized probability. Therefore, the SPN rooted at sum node $i$ can be viewed as a mixture model, where each child $j$ is a mixture component with distribution $S_j(\mb{x})$.

\subsubsection{Inference}\label{inf}
We can perform marginal inference and MPE inference with a valid SPN in time linear to its size. This uses the same idea as Darwiche's derivation of partial differentiations in arithmetic circuits \cite{darwiche2003differential}.

Given an SPN $S$ and an arbitrary intermediate node $i$, let $S_i(\mb{x})$ be the value of the node on state $\mb{x}$. Let $Pa(i)$ be the parent nodes of $i$. Then,

\begin{equation}
\label{psx}
    \frac{\partial S(\mb{x})}{\partial S_i(\mb{x})}= \sum_{k\in Pa(i)}\frac{\partial S(\mb{x})}{\partial S_k(\mb{x})} \frac{\partial S_k(\mb{x})}{\partial S_i(\mb{x})}
\end{equation}
If node $i$ is a product node, $\frac{\partial S_k(\mb{x})}{\partial S_i(\mb{x})}=w_{ki}$. If node $i$ is a sum node, $\frac{\partial S_k(\mb{x})}{\partial S_i(\mb{x})}=\prod_{l\in Ch_{-i}(k)}S_l(\mb{x})$. We can compute $\frac{\partial S(\mb{x})}{\partial S_k(\mb{x})}$ by first going upwards from leaves to root then going downwards from root to node $i$. 

\paragraph{Marginal inference}\label{if:mg}
Suppose node $i$ is an indicator $x_p^a$. Then, as derived in \cite{darwiche2003differential},

\begin{equation}
P(X_p=a|\mb{e})=\frac{1}{S(\mb{e})}\frac{\partial S(\mb{e})}{\partial S_i(\mb{e})}\propto\frac{\partial S(\mb{e})}{\partial S_i(\mb{e})}
\end{equation}
We can also infer the marginal of hidden variables. Suppose node $i$ is a sum node which marginalizes out a hidden variable $Y_i$. For child $j$ branching out from $i$, we can consider the value of indicator $[Y_i=j] = S_j(\mb{e})$. This holds because the indicators are actually real-valued \cite{darwiche2003differential}\cite{peharz2015theoretical}, and this is important when we take the derivative with respect to an indicator \cite{peharz2015theoretical}. Thus, we use the partial differentials derived by Darwiche \cite{darwiche2003differential} to obtain the following. Note that by definition, $\mb{E}$ and $\mb{Y}$ must be disjoint.
\begin{equation}\label{eq:pyi}
P(Y_i=j, \mb{e})=\frac{\partial S(\mb{e})}{\partial S_j(\mb{e})}
\end{equation}
\begin{equation}
  \begin{aligned}
    P(Y_i=j|\mb{e})= \frac{P(Y_i=j, \mb{e})}{P(\mb{e})} &= \frac{1}{S(\mb{e})}\frac{\partial S(\mb{e})}{\partial S_j(\mb{e})}\\
    &= \frac{1}{S(\mb{e})}\frac{\partial S(\mb{e})}{\partial S_i(\mb{e})}\frac{\partial S_i(\mb{e})}{\partial S_j(\mb{e})}\\
    &=\frac{w_{ij}}{S(\mb{e})}\frac{\partial S(\mb{e})}{\partial S_i(\mb{e})}\\
    &\propto w_{ij}\frac{\partial S(\mb{e})}{\partial S_i(\mb{e})}
  \end{aligned}
\end{equation}

\paragraph{MPE inference}\label{if:map}
The task of MPE inference is to find the most likely assignment of unobserved variables given the evidence. Suppose we have evidence $\mb{e}$ for a set of observed variables $\mb{E}$, we have hidden variables $\mb{Y}$, and we have the unobserved input variables $\mb{X}=\mathcal{X}-\mb{E}$. We are computing $\argmax_{\mb{x},\mb{y}}P(\mb{x},\mb{y}|\mb{e})$.

This can be done by replacing each sum operation with a maximization operation, such that $S_i(\mb{e})=\max_{j}w_{ij}S_j(\mb{e})$. First, we compute $S_i(\mb{e})$ for every node from bottom to top. Next, we traverse recursively from the root to the leaves: At a sum node, we pick its child that led to the value of that sum node; At a product node, we select all of its children for traversal. 

If $S$ is consistent, it is guaranteed that there is no conflicting values among the children of a product node, which means that the MPE result is a valid instantiation of the variables. In this case, the obtained instantiation $\hat{\mb{x}}$ has the maximum value of $S(\hat{\mb{x}}|\mb{e})$ \cite{poon2011sum}.

\subsubsection{Learning Parameters}

\paragraph{Gradient descent and backpropagation} Suppose $S$ is a valid SPN over the set of variables $\mathcal{X}$, and suppose $w_{ij}$ represents the weight of the edge from sum node $i$ to its $j$-th child. Suppose we have observations of states $\mb{x}=\{\mb{x}_1,\cdots,\mb{x}_n\}$. We aim to estimate the weights $\mb{w}$ by maximizing the log-likelihood function:
\begin{equation}
l(\mb{w}|\mb{x})=\sum_{p=1}^{n}logP_{\mb{w}}(\mb{x}_p)=\sum_{p=1}^{n}log\Big(\frac{S(\mb{x}_p)}{Z}\Big)
\end{equation}
where $Z$ is the partition function that equals to $S(1\cdots1)$. The second equal sign holds because $S$ is valid. Now we take the gradient with respect to a single weight $w_{ij}$:
\begin{equation}
\label{dl}
\frac{\partial l(\mb{w}|\mb{x})}{\partial w_{ij}}=\sum_{p=1}^{n}\frac{\partial}{\partial w_{ij}}log\Big(\frac{S(\mb{x}_p)}{Z}\Big)=\sum_{p=1}^{n}\frac{1}{S(\mb{x}_p)}\frac{\partial S(\mb{x}_p)}{\partial w_{ij}}-\sum_{p=1}^{n}\frac{1}{Z}\frac{\partial Z}{\partial w_{ij}}
\end{equation}
The value of the sum node $i$ is $S_i(\mb{x}_p)=\sum_{j\in Ch(i)}w_{ij}S_j(\mb{x}_p)$. We can thus use the chain rule to obtain:
\begin{equation}
\frac{\partial S(\mb{x}_p)}{\partial w_{ij}} = \frac{\partial S(\mb{x}_p)}{\partial S_i(\mb{x}_p)}\frac{\partial S_i(\mb{x}_p)}{\partial w_{ij}}=\frac{\partial S(\mb{x}_p)}{\partial S_i(\mb{x}_p)}S_j(\mb{x}_p)
\end{equation}
Using (\ref{psx}), we can compute $\partial S(\mb{x}_p)/\partial S_i(\mb{x}_p)$ by first computing the partial derivative with respect to the parents of $i$. Computing $\partial Z/\partial w_{ij}$ follows similar derivation. This is naturally a backpropagation process. Alternatively, we can ensure that $S(\mb{x}_p)$ to be a normalized probability by renormalizing the weights after each full update, so that we can discard the $Z$ in the derivation.

\paragraph{Hard EM} As discussed previously, a sub-SPN rooted at sum node $i$ in an SPN $S$ can be viewed as a mixture model. 
%% We have discussed hidden variables the sum node $i$ implicitly marginalizes out when $\sum_{j\in Ch(i)}w_{ij}=1$. To estimate the model parameters $\mb{w}$, EM is a natural choice in this situatinoro. I attempt to derive the EM for SPN weight update in appendix \ref{apdx:spn}.
%\paragraph{Hard EM}
Poon and Domingos found that their EM method does not work well in practice, so they opt to use MPE inference instead of the marginal inference in the E step, and changes the M step accordingly. The algorithm can be described as follows.

\begin{enumerate}
    \item[(1)] In E step, compute MPE inference:
        \begin{equation}
        j^*=\argmax_j P(\mb{x}_p, Y_i=j)=\argmax_j\frac{\partial S(\mb{x}_p)}{\partial S_j(\mb{x}_p)}=\argmax_j w_{ij}\frac{\partial S(\mb{x}_p)}{\partial S_i(\mb{x}_p)}
        \end{equation}
        A count is kept for every child of the node $i$ to record the number of times the index of that child equals to $j^*$.
        
    \item[(2)] In M step, the count of child $j^*$ increments by one.
\end{enumerate}

\subsubsection{Learning Structure}

\paragraph{LearnSPN}
The most commonly used structure learning algorithm is LearnSPN \cite{gens2013learning}. This algorithm recursively divides up the training data either by rows (instances) or columns (features). A product node is added when the variables can be partitioned into approximately independent subsets. If this is not the case, the instances are partitioned into similar subsets, and sum node is added on top of the subsets. This algorithm requires a large amount of training data to yield good partition results. This is not very feasible in our problem because there is very little data for the topological maps and it is difficult to acquire a massive amount of such data.

\begin{figure}[!htb]
    \centering
    \captionsetup{width=.8\linewidth}
    \includegraphics[scale=0.6]{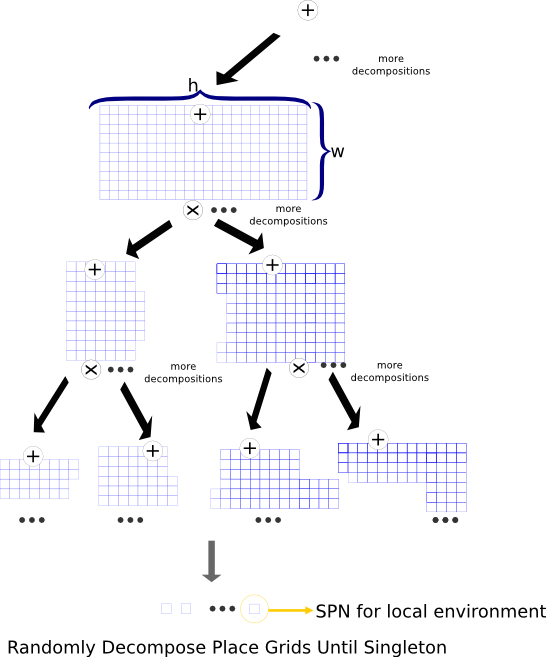}
    \caption{Process of constructing a valid SPN using the random-decomposition algorithm. Each sum node models a mixture of different decompositions of the variables in its scope. Each product node unions the scopes its children by multiplying them together. The leaf (singleton) variable, in our context, is a place which can be modeled by an SPN in DGSM \cite{pronobis2016learning}.}
    \label{fig:grid_learn}
\end{figure}

\paragraph{Random-Decomposition}\label{sc:rand}
We use a different approach in learning the SPN structure, similar to the one used in \cite{pronobis2016learning}. This algorithm is visualized in figure \ref{fig:grid_learn}. Given a set of variables $\mathcal{X}$, we decompose it into random subsets multiple times. A sum node is used to model a mixture of these decompositions, summing up the output of SPNs built upon each of these decompositions. This process is done recursively for each subset until singleton. Product nodes combine the output of the SPNs on each subset to union their scopes. Weights are shared for each sum node at the same level, and edges with zero weight are pruned once the parameter learning is finished.

\subsection{Deep Affordance Spatial Hierarchy}\label{section:dash}

Pronobis et.~al.~\cite{pronobis2017deep} proposed a hierarchical spatial representation named Deep Affordance Spatial Hierarchy (DASH). In their work, they also discussed the properties of a desired representation. We summarize these properties as follows: 

\begin{enumerate}[label=(\arabic*)]
    \item Minimal;
    \item Hierarchical. It allows long-term, high-level, global planning to break down to short-term, lower-level local planning;
    \item Correlating abstraction with information persistence. In other words, more dynamic means less persistent, which requires higher level of abstraction;
    \item Probabilistic;
    \item Representing the unknowns, that is, it allows the representation of default knowledge;
\end{enumerate}

\noindent Guided by the above principles, the authors presented DASH, which is aimed to be used by deep learning frameworks. DASH consists of four layers.
\begin{enumerate}[label=(\arabic*)]
    \item Perceptual layer represents the robot's raw sensor readings. In the paper, the authors used occupancy grid to represent the laser-range observations, centered at the robot's position.
    \item Peripersonal layer represents objects, obstacles, and landmarks in the space immediately surrounding the robot. The authors used a polar occupancy grid representation which is a simplification of this layer's information. 
    \item Topological layer represents the graphical structure of the map that connects places together. It maintains information of the places, such as place categories and views at different angles, and the connectivity between places. It also contains placeholders. In the paper, a topological map is constructed from first sampling places in a window centered at the robot, according to the probability distribution $P(E|G)$, where $E$ is the variable for existance of places, and $G$ is the occupancy grid provided by the perceptual layer. Then, according to navigation affordance, places reachable from the robot are added. The affordance is determined by A* search over the potential field $P(E|G)$.  More details on topological map generation is discussed in the next section.
    \item Semantic Layer represents human concepts about the places. It can be visualized as a concept map that specifies the relations between concepts. The knowledge represented in this layer can be used for complex human-robot interaction tasks. In the paper, the authors implemented a probabilistic relational data structure that mimics a concept map.
\end{enumerate}
When the robot is handed with a DASH, ideally it is supposed to be able to navigate around and complete tasks that involve interaction with humans. And the robot should be able to explore the environment and build up the topological layer on its own. However, on its own, it does not have the ability to model uncertainty in the environment, and infer semantics of placeholders or other latent variables. To enable this, the authors used DGSM (Deep Generative Spatial Model) \cite{pronobis2016learning} to represent the default knowledge about the environment. Default knowledge can be understood as the knowledge that the system assumes before it is provided explicitly. In the context of the paper, default knowledge means place semantics that the model infers for all places (including placeholders) based on its training data. In this thesis, we also focus on using place categories as our semantic information. While DGSM learns a local version of the DASH spatial knowledge, we enable the learning of the topological layer in this thesis. Eventually, we look to connect the two so that the entire spatial knowledge hierarchy is represented.

\subsection{Topological Maps}\label{section:topomap}

\begin{figure}[!htb]
    \centering
    \captionsetup{width=.8\linewidth}
    \includegraphics[scale=0.5]{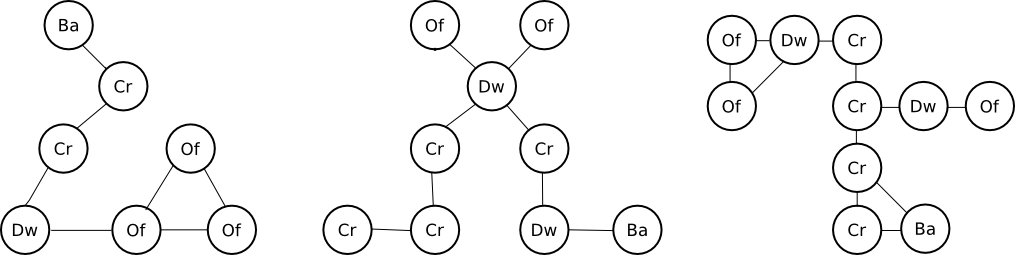}
    \caption{Simplified example topological maps. The text on a node indicates the category of the place (Ba = bathroom, Cr = corridor, Dw = doorway, Of = office).}
    \label{fig:topomap}
\end{figure}

A topological map is a graph that captures important places in the map as well as their connectivity. In the context of robotics, the nodes in this graph are the places accessible by the robot, and edges indicate navigation affordance between places, i.e. the feasibility of navigating from one place to another. Figure \ref{fig:topomap} shows examples of simplified topological maps.

As discussed previously, topological map is used as a layer to capture the geometry of the map and connectivity between places in the DASH. The authors of DASH \cite{pronobis2017deep} also describe a mapping algorithm that builds a topological map. As the robot explores an environment, the mapping algorithm expands the topological map by adding \textit{placeholders} \cite{pronobis2010representing} as nodes, according to a probability formulation as follows
\begin{equation}\label{eq:p_eg}
  P(E|G) = \frac{1}{Z}\prod_i\phi_I(E_i)\phi_N(E_i, \mathcal{E})
\end{equation}
where $P(E|G)$ is the probability of placeholder locations $E$ given robot-centered occupancy grid $G$ which represents laser-range observations. The potential function $\phi_I(E_i)$ models the existence of placeholder at location $i$, denoted as $E_i\in\{0,1\}$. For details about how $\phi_I(E_i)$ is defined, refer to the original paper \cite{pronobis2017deep}.. The potential function $\phi_N(E_i, \mathcal{E})$ models the probability of placeholder $E_i$ in the neighborhood of the set of existing places $\mathcal{E}$. It is defined as
\begin{equation}\label{eq:phi_n}
  \phi_N(E_i, \mathcal{E}) = \sum_{p\in\mathcal{E}}\exp\Big(-\frac{(d(i,p)-d_n)^2}{2\sigma^2}\Big)
\end{equation}
where $d(i,p)$ is the distance between location $i$ and place $p$. The key point to note here is that $\phi_N(E_i, \mathcal{E})$ promotes places that are of certain distance $d_n$ apart from existing places. This fact is important when we describe the place grid method in section \ref{section:grid}.

Using this method, in addition to capturing topological relations between places, the overall geometric structure of the full map is also preserved in the topological map, since each node has its coordinates on the metric map. For example, corridor nodes usually form a long and straight line, room nodes are usually cluttered, and doorway nodes usually have edges coming out from two opposite sides. We hope to learn these general geometric properties of place categories as well.

\subsubsection{Challenges}
There are several challenges in modeling topological maps due to their scale and variability. A full map of a workspace may contain dozens of nodes and hundreds of edges. The number of places in one map likely differs from that in another. Different topological maps may represent environments of different dimensions. The structure of places in one map may appear very differently compared to another. Finally, because our topological maps can encode geometric structure as described above, even for similar environments, different topological maps may be rotated differently, since the underlying metric maps, built by SLAM (Simultaneous Localization and Mapping), may be rotated differently as a whole.

\begin{figure}[!htb]
  \centering
  \begin{subfigure}{.4\textwidth}
  \centering
  \includegraphics[width=.5\linewidth]{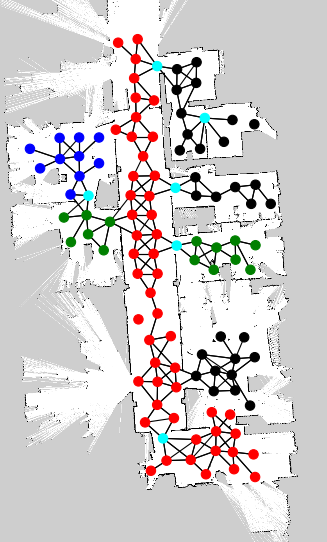}
  \label{fig:sfig1}
\end{subfigure}%
\begin{subfigure}{.4\textwidth}
  \centering
  \includegraphics[width=.7\linewidth]{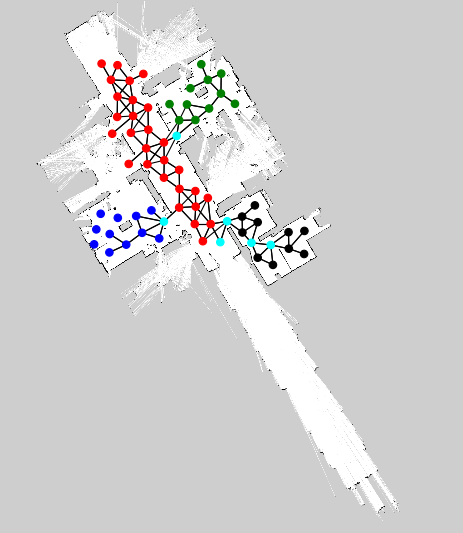}
  \label{fig:sfig2}
\end{subfigure}
\captionsetup{width=.8\linewidth}
\caption{Example topological maps. Colors indicate place categories. Note that the underlying metric maps have different rotations.}
\end{figure}

%%%%%%%%%%%%%%%%%%%%%%%%%%%%%%
% Proposed Solutions
%%%%%%%%%%%%%%%%%%%%%%%%%%%%%%
%\newpage
\section{Problem Statement \& Proposed Solutions}\label{section:sol}
\subsection{Problem Statement}\label{section:problem}

Our problem is the following. Given datasets of topological maps with unknown number of nodes per map, we would like to train Sum-Product Networks such that the resulting model is able to demonstrate its understanding of the spatial relations of categories.

It can demonstrate this ability in several ways. For example, if we provide a topological map instance with missing semantic information, the model should reasonably predict what the missing information is. In addition, if we provide a topological map instance with strange category arragements (e.g. corridor nodes labeled office), the model should be able to detect this novelty. In section \ref{section:experiment}, we describe experiments conducted to test out this ability.

\subsection{Place Grid Method}\label{section:grid}
To address the challenge that topological maps are of different sizes and shapes, we present a simple idea which is to map them to fixed-size grids. A grid can be considered as a generic representation of topological maps. It approximates the topological map, but preserves most of the adjacency relations between places. It is straightforward to use SPN to model this grid. Using the algorithm in \ref{sc:rand}, we can train an SPN for the grid, which then indirectly models topological maps as a whole. 

\subsubsection{From Topological Map to Place Grid}

In this section, we first define a \textit{place} in a topological map, then define a \textit{place grid}, and then describe in detail how can a topological map be mapped to a place grid. Our definition of a place follows from the polar occupancy grid representation for local environment described by Pronobis and Rao in \cite{pronobis2016learning}.

\begin{definition}[Place]
A \textit{place} $p$ in a topological map is a three-tuple $(L, V, a)$ where $L$ is the location of $p$ with respect to the topological map's coordinate system, $V$ is the set of views in the polar occupancy grid, and $a$ is the label for place category.
\end{definition}

\begin{figure}[!htb]
    \centering
    \captionsetup{width=.5\linewidth}
    \includegraphics[scale=0.15]{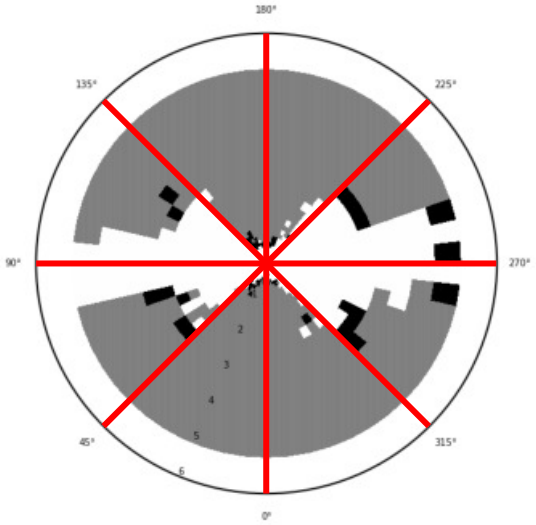}
    \caption{Example place, a polar occupancy grid divided into 8 views, labeled as doorway. (Image adapted from \cite{pronobis2016learning})}
    \label{fig:place}
\end{figure}

\noindent An example polar occupancy grid of a place is shown in figure \ref{fig:place}.

\begin{definition}[Place Grid]
A place grid $G$ with resolution $r$ (meter per cell) is an $m\times n$ matrix where each cell contains a place.
\end{definition}

Given a 2D place grid $G$ with resolution $r$, we can map a place $p$ in topological map $M$ with $L=(p_x^M, p_y^M)$ to a cell in $G$ at $(p_x^G, p_y^G)$ simply by
\begin{equation}
  (p_x^G, p_y^G) = \Big(\floor{\frac{p_x^M-p_{x_{min}}^M}{r}}, \floor{\frac{p_y^M-p_{y_{min}}^M}{r}}\Big)
\end{equation}
where $p_{x_{min}}^M$ is the $x$ coordinate for the place with minimum $x$ coordinate (similar goes for $p_{x_{min}}^M$).

To avoid gaps between places mapped to the grid which disturb adjacency relations between places, we can map the topological map to a place grid with resolution at most $d_n$. (Recall from equation (\ref{eq:phi_n}) that places in a topological map are separated by a predefined distance $d_n$.)\label{sc:avoidgaps}

To deal with the situation where multiple places map to the same cell, we can do either of the following:
\begin{itemize}
\item Pick the place with the highest value of $\phi_I(E_p)$, as shown in equation (\ref{eq:p_eg}). 
\item Create a hybrid place by sampling views from these places with probability according to $\phi_I(E_p)$.
\end{itemize}

We conduct experiments to test this place grid method (section \ref{section:experiment}). Because the training and validation topological map instances of our experiment are built from one single building, these instances do not vary much in rotation. In general, however, we need to deal with the case where we have no assumption of from where the topological maps are generated. Next, we provide a solution in theory to deal with topological maps with different rotations, in the context of the place grid method.

\subsubsection{Modeling a Place Grid Using Rotation-Invariant SPN}

There are multiple ways to deal with the rotation of topological maps. We could rotate all the topological maps before hand, based on the prior knowledge that human environments typically have right angle corners and straight walls. Another way is to incorporate rotation-invariance into the SPN that models the place grid. Here, we present the theory of constructing a rotation-invariant SPN based on a place grid.

\paragraph{Rotation-invariant SPN} We consider $k$ number of major directions (e.g. 8). At the top level of the SPN, we add a max node with $k$ children of sum nodes where each sum node models the place grid for a certain orientation. Suppose one child $A$ is a grid resulted from rotating another child $B$ by angle $\theta$. Then, a cell $(x_B, y_B)$ in $B$ corresponds with (has the same value as) a cell $(x_A, y_A)$ in $A$ by the following transform:

\begin{equation}
\begin{bmatrix}
x_B \\
y_B \\
\end{bmatrix}
=
\begin{bmatrix}
\floor{cos(\theta)x_A+sin(\theta)y_A}\\
\floor{-sin(\theta)x_A+cos(\theta)y_A}\\
\end{bmatrix}
\end{equation}

With the above relation, we are not adding more inputs to the SPN. We simply connect a cell in the grid with the corresponding input for the place. There is a case where this relation causes $(x_B, y_B)$ to be out of bound, which means that there are some cells in the ``rotated'' grid $B$ that have no corresponding cells in $A$. We simply ignore the out of bound cells, and feed 1 as input for those cells that have no correspondence. One way to reduce the effect of this on the trained SPN is by using a place grid with large dimension where there are loops of cells that are supposed to be empty. The point is that these cells do not really affect the SPN's representation of the useful parts in the place grid where mapping happens.

%% \begin{figure}[!htb]
%%     \centering
%%     \captionsetup{width=.8\linewidth}
%%     \includegraphics[scale=0.6]{images/grid_method.png}
%%     \caption{High-level process of the Place Grid Method. We have several place grids with different resolutions, each is modeled by an SPN using the random-decomposition algorithm. The topological maps are mapped to the grids to provide training examples for the SPNs.}
%%     \label{fig:topomap}
%% \end{figure}

%% At this point, we can certainly map any topological map to grids of different resolutions, and use an SPN to model each. Lower resolution means the grid represents relations of places at a more general level. This effectively results in an ensemble of SPNs (see figure \ref{fig:topomap})). If a query concerns, for example, what category should a place at a location in the topological map be given its surroundings, we can have the ensemble of SPNs to each come up with an answer and combine them in a certain way (e.g. weighted voting) to produce the final answer. 

\subsection{Template-Based Method}\label{section:tmpl}

\subsubsection{Motivation}
One shortcoming of the place grid method is its rely on a predefined size and resolution of the grid; choosing the right dimensions and resolution is a difficult task and usually the chosen parameters does not generalize well for other unseen topological map instances. Besides, the grid itself is in the middleground between being accurate and approximate: Althougth the grid is metric, the cells do not accurately represent the location of places. This may affect the quality of inference.

To some extent, it is unnecessary to have an accurate absolute representation of the topological map, because (1) it is hard to maintain this accuracy given the variability of topological maps, and (2) topological maps are usually used in conjunction with metric grid maps for robot planning and navigation, serving as a reference for high-level planning. Therefore, we propose a template-based relative representation of the topological map, which models some certain basic structure of a graph, and expands as needed. As discussed before, the findings by  Aydemir et.~al.~\cite{aydemir2012can} support the idea of using a subgraph of small size as the template for the entire topological map.

\begin{figure}[!htb]
    \centering
    \captionsetup{width=.8\linewidth}
    \includegraphics[scale=0.5]{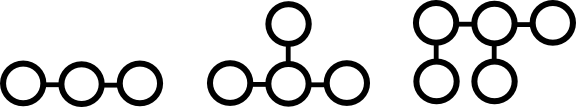}
    \caption{Some simple basic structures.}
    \label{fig:basic_struct}
\end{figure}

\begin{figure}[!htb]
    \centering
    \captionsetup{width=.8\linewidth}
    \includegraphics[scale=0.4]{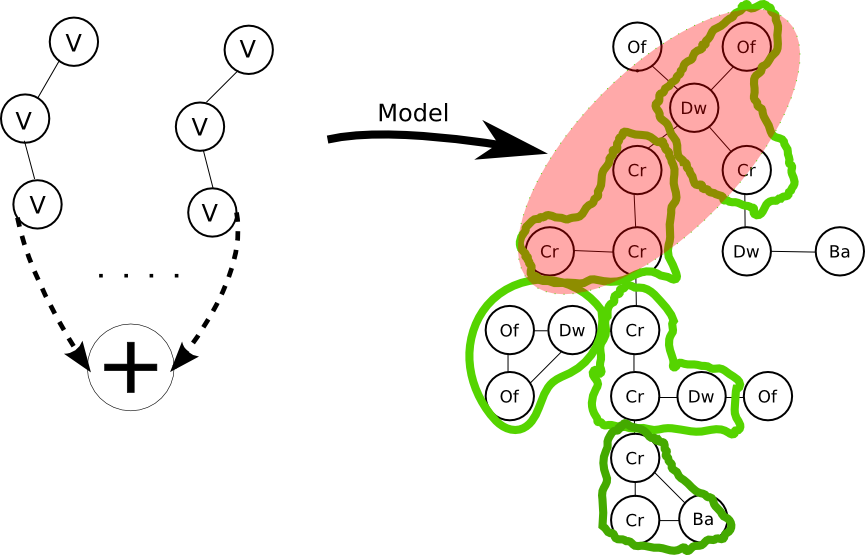}
    \caption{Illustration of the Template-Based Method. On the left side, the three node template is the basic structure $\mathcal{B}$, and the pair of these template form a hierarchy of templates, on which the SPN is constructed. On the right side, a graph instance can be partitioned into a supergraph using $\mathcal{B}$, and the SPN essentially models the red region on the graph.}
    \label{fig:tmpmod}
\end{figure}

\subsubsection{Building Instance-Specific SPN for Graphs from Templates}\label{section:buildtmplspn}
Consider a basic graph structure $\mathcal{B}$, such as the three-node structure in figure \ref{fig:tmpmod}. The only constraint that this structure has is the connectivity of the places; How the views are connected is not constrained. Therefore, this basic structure is rotation-insensitive. We can partition a given graph instance using $\mathcal{B}$. Note that $\mathcal{B}$ should be a very simple structure. Examples of such basic structure are shown in figure \ref{fig:basic_struct}. One reason is that a simple structure likely occurs very often in the topological map. Another reason is that finding if a graph is a subgraph of another (subgraph isomorphism problem) is NP-complete. Linear time solution exists but only works for planar graphs \cite{eppstein1995subgraph}, and a topological map is not necessarily planar.

\begin{figure}[!htb]
    \centering
    \captionsetup{width=.8\linewidth}
    \includegraphics[scale=0.5]{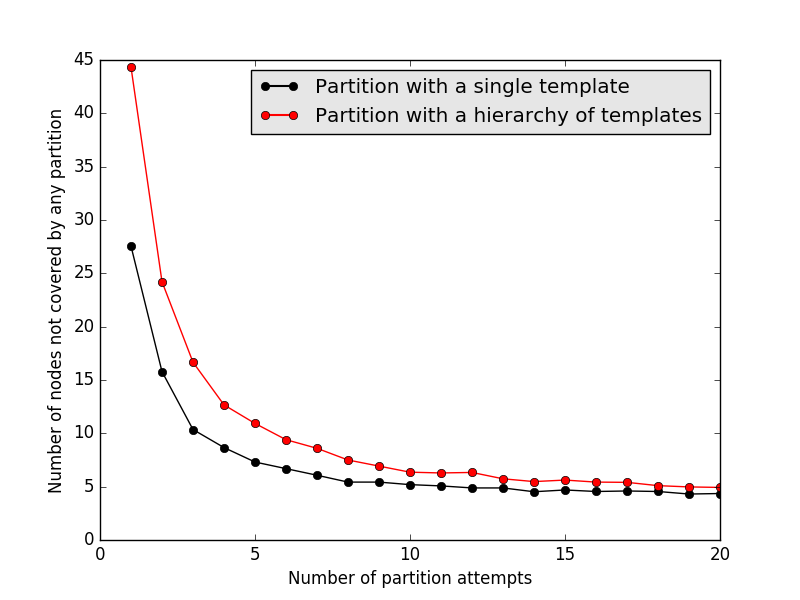}
    \caption{The relation between number of partition attempts and the number of nodes uncovered by any of these attempts.}
    \label{fig:part_drop}
\end{figure}

Algorithm \ref{alg:part_graph} is a greedy algorithm to partition the graph when $\mathcal{B}$ is very simple. In this algorithm, \textbf{match\_graph} is assumed to be a function that is specific to $\mathcal{B}$; it could be hard-coded. The output of this algorithm is a graph $H$ where nodes are the basic structures and edges indicate their connectivity (see the green groups in figure \ref{fig:basic_struct}). It is likely that the output of this algorithm is a graph that does not cover all nodes from the input graph. Yet, as shown in figure \ref{fig:part_drop}, the number of uncovered nodes drops rapidly as we increase the number of partition attempts, that is, the number of times we run this algorithm. With this, we know that we can cover the topological relations in the graph better if we partition the graph multiple times, both in training and testing.

\begin{algorithm}
    \caption{Graph Partition by a Basic Structure}
    \label{alg:part_graph}
    \SetKwInOut{Input}{Input}
    \SetKwInOut{Output}{Output}

    \underline{function Graph-Partition} $(G, \mathcal{B})$\;
    \Input{A graph $G=(V_G,E_G)$ and a graph $\mathcal{B}$ representing a basic structure.}
    \Output{A graph $H=(V_H,E_H)$ resulting from the partition of $G$ using $\mathcal{B}$.}
    $V_{available}\gets V_G$\;
    $M \gets$ empty map\;
    \While{$V_{available} \neq \varnothing$}{
      $v\gets$ draw from $V_{available}$ randomly\;
      $V_{used}\gets \mathcal{B}.$\textbf{match\_graph}$(G, v)$\;
      \If{$V_{used}\neq \varnothing$}{
        $V_{available}\gets V_{available} - V_{used}$\;
        $u\gets$ a node that represents $V_{used}$\;
        $V_H\gets V_H \cup \{u\}$\;
        \ForEach{$v'\in V_{used} \cup \{v\}$}{
          $E_H\gets E_H \cup (u, M.\textbf{get}(v'))$\;
          $M.$\textbf{put}$(v', u)$\;
        }
      }
      $V_{available} = V_{available} - \{v\}$\;
    }
    \KwRet{$(V_H, E_H)$}
\end{algorithm}

\begin{figure}[!htb]
  \centering
  \begin{subfigure}{.3\textwidth}
  \centering
  \includegraphics[width=.8\linewidth]{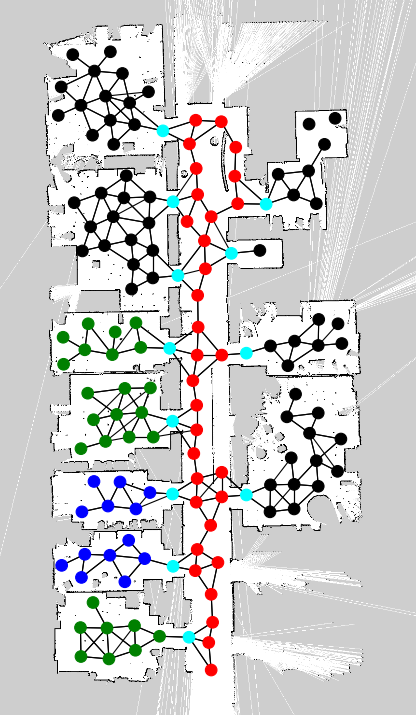}
  \caption{}
  \label{fig:orig_topo}
\end{subfigure}%
\begin{subfigure}{.3\textwidth}
  \centering
  \includegraphics[width=.77\linewidth]{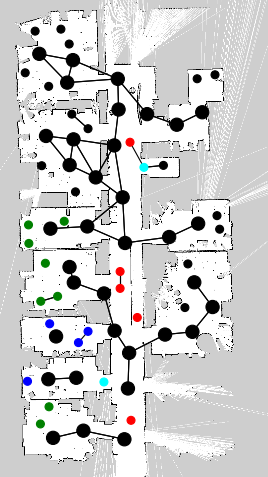}
  \caption{}
  \label{fig:part_tmpl}
\end{subfigure}
\begin{subfigure}{.3\textwidth}
  \centering
  \includegraphics[width=.8\linewidth]{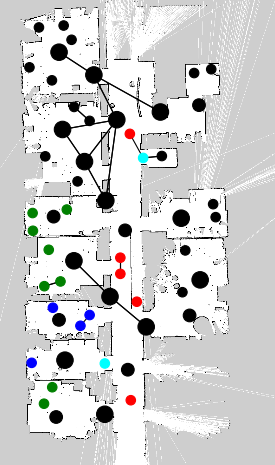}
  \caption{}
  \label{fig:part_hier}
\end{subfigure}
\captionsetup{width=.8\linewidth}
\caption{ Sequence of partitioning a topological map graph using a hierarchy of templates. (a) shows a topological map. (b) shows a graph produced by partitioning the topological map using the three-node template, as in figure \ref{fig:basic_struct}. (c) shows a graph produced by partitioning the graph in (b) using the pair template (two nodes). The size of node increases as the number of nodes covered in the original topological map increases; In (b) each largest node covers 3 nodes in (a). In (c) each largest node covers 6 nodes in (a). The four non-black colors (blue, cyan, green, red) indicate place categories (small office, doorway, large office, corridor). Black indicates the node has no associated place category. Note that there are some nodes left-out uncovered by one or all of the partitioning.}
\end{figure}

Now that we have a basic structure as our template, we can consider constructing a \textit{hierarchy} of the templates. Consider another simple graph structure $\mathcal{S}$ that consists of $\mathcal{B}$. For example, $\mathcal{S}$ is a pair of $\mathcal{B}$ connected as shown in the red region of the topological map in figure \ref{fig:tmpmod}. We can again partition the graph $H$ by $\mathcal{S}$ using the same algorithm. Therefore, the given graph instance can be partitioned into a graph with nodes that represent $\mathcal{S}$'s.

We can think of $\mathcal{B}$ as the template, and $P(\mathcal{S})$ as the joint distribution of the templates defined in $\mathcal{S}$. The probability $P(\mathcal{S})$ is the \textit{expansion model} of the template that dictates how a template expands to form a graph. Unlike other template-based models such as hidden Markov Models (HMMs), we don't have to use a transition model between the templates, because there is no notion of "order" between the places in a topological map.

\begin{figure}[!htb]
    \centering
    \captionsetup{width=.8\linewidth}
    \includegraphics[scale=0.5]{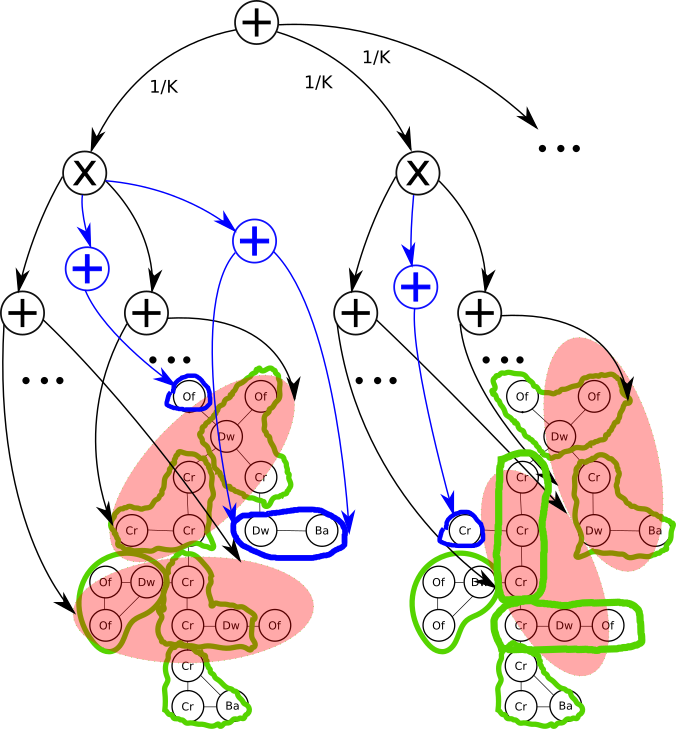}
    \caption{Full SPN constructed for an instance of topological map. Each child of the root node indicates one partition attempt of the topological map by a hierarchy of templates (pair of three-node structure). There is total of $K$ partition attempts, so the weight for each child is $1/K$. Each black-colored sum node (besides the root) indicates the root of the SPN that spans a hierarchy of templates, and these SPNs share the same weight values. The blue-colored sum node indicates the root of the SPN that spans the uncovered nodes during the partitioning. Note that for clarity, SPNs that span the nodes grouped by green but not covered by red shade are not shown.}
    \label{fig:full_spn_illus}
\end{figure}

We can construct an SPN for the hierarchy of templates using the random-decomposition method. The resulting SPN is likely small. We can perform inference at the scale of the entire topology map by constructing an instance-specific SPN: For a given test case, we partition the topological map using the same structures $\mathcal{B}$ and $\mathcal{S}$. For each hierarchical template structure resulted from the partitioning, we construct a sub-SPN that has the exact same structure and weights as the learned hierarchical template SPN. We combine these sub-SPNs together simply by a product node.  We do this partition multiple times to capture as much combination of nodes that form the template structures as possible. We use a sum node to sum up the product nodes that represent the partition attempts. The resulting instance-specific SPN for a topological map is illustrated in figure \ref{fig:full_spn_illus}.

Note that there are uncovered nodes after partitioning (see figure \ref{fig:tmpmod}). Each product node that represents a partition attempt needs to cover these nodes as well, otherwise the sum node at the root would have children with different scopes, leading to invalid SPN. These uncovered nodes cannot form any more template structures. 
Therefore, we can train separate SPNs to cover them. For example, if we use the three node structure as the template for partitioning the graph, the uncovered nodes would either be single nodes or pairs of nodes. We train an SPN for a trivial single node template, and an SPN for a pair template. We can them use these template SPNs when constructing the full SPN for the entire topological map.

% We recursively partition the full global map by B and S in alternating fashion, and the weights for the SPN representing B and S are all shared.

%%%%%%%%%%%%%%%%%%%%%%%%%%%%%%
% Experiments
%%%%%%%%%%%%%%%%%%%%%%%%%%%%%%
\newpage
\section{Experiments}\label{section:experiment}
\subsection{Data Collection}\label{section:data}

In support of the larger goal of our research to learn the Deep Affordance Spatial Hierarchy (DASH), we collected a large dataset of fully-annotated sensory (RGB, laser rangefinder readings, odometry), localization, and virtual scans, a low-level representation of the environment immediate to the robot. For this thesis, we utilized this dataset to generate topological maps with nodes labeled by place categories.

The dataset consists of 100 sequences and total of 11 different floor plans in three buildings in three different locations: Stockholm, Sweden, Freiburg, Germany, and Saarbrucken, Germany. Summary of the sequences is shown in table \ref{table:dataset}. For each sequence, among the types of data contained,  there is a corresponding canonical map and annotated floor plan, on which the annotated localized poses are generated using adaptive Monte-Carlo Localization (AMCL) with frequency of 30~Hz. We use the canonical map and the localized poses to generate topological maps, one for each sequence. We used $1.0$m as the desired separation between place nodes when generating topological maps. We trained and tested our methods mostly on the Stockholm sequences as the building in Stockholm has much larger number of rooms and bigger in scale.

\begin{table}[h!]
\centering
 \begin{tabular}{|l|c|c|c|} 
 \hline
  & Stockholm & Freiburg & Saarbrucken \\ [0.5ex] 
 \hline
 \# sequences & 42 & 26 & 32 \\ 
 \hline
 \# floor plans & 4 & 3 & 4 \\
 \hline
 Avg. \# topological map nodes per sequence & 103.62 & 66.12 & --\footnote{We did not generate topological maps for the Saarbrucken sequences.} \\
 \hline
 Avg. \# topological map edges per sequence & 159.90 & 104.86 & -- \\
 \hline
 \end{tabular}
 \caption{Details of the dataset}
 \label{table:dataset}
\end{table}

\subsection{Methodology}\label{section:method}
\subsubsection{Software}
We used LibSPN \cite{pronobis2016learning}, a Python library that implements learning and inference with SPNs on GPUs. This library provides the functionality to generate a dense SPN structure using the random-decomposition method as described in section \ref{sc:rand}. We use this functinoality to generate SPN on top of a place grid as well as on templates. We implemented ourselves weight-sharing and instance-specific full SPN construction as needed for the template-based method. 

\subsubsection{Experiments for Place Grid Method}\label{section:exp-grid}

We converted topological maps from Stockholm sequences into place grids. We used place grids of size of 15 rows by 10 columns with resolution of 1.5m per cell. The resolution is lower than the 1.0m of desired place separation in topological map to ensure the connectivity of places when mapped to the grid, as suggested in section \ref{sc:avoidgaps}. Since the size of our dataset is not very large, we lowered the number of place categories to four: doorway, corridor, small office, and large office. so that we can better assess the learning of structure and dependency. Note that for other categories such as kitchen and meeting room, we do not distinguish them from the \textit{unknown} place category.

To verify the consistency of the place grid method,  we used cross-validation by training the network on three floors and test it on the other one, for each combination of the four floors in the Stockholm dataset. The training procedure's likelihood curve is shown in \ref{fig:likelihood}. As we can see, the standard deviation of the likelihood values is small, which suggests that training on either three of the four floors results in a very similar model.

We did not implement the rotation-invariant SPN when testing the place grid method, as the canonical maps in the Stockholm dataset have the same orientation, and the topological map data examples are aligned well with each other. In the future, we plan to let the SPN learn the topological maps from two datasets (e.g. Stockholm and Freiburg) and test it on another (e.g. Saarbrucken), and implement the rotation-invariant SPN.

We test the method by using it for two main types of tasks, \textit{map completion} and \textit{novel map structure detection}. For map completion, we use the SPN to do MPE inference for the full map and the occluded part of the map. Also, we randomly occlude cells in the place grid that may or may not have been mapped directly from the topological maps, and test if the network can infer their values correctly. For novel structure detection, we construct test examples of topological maps by swapping the labels of pairs of classes, such as doorway and corridor, large office and small office, and small office and corridor.

\subsubsection{Experiments for Template-Based Method}\label{section:exp-tmpl}

As described in section \ref{section:tmpl}, during training, we train an SPN for a single template, or a hierarchy of templates. Therefore, we created a dataset where each example is an assignment of the template variables. The dataset is created from both Stockholm sequences and Freiburg sequences, because the topological relations of place categories are generally the same across different environments
\cite{aydemir2012can}. We do this by partitioning the available topological maps using the templates. The partition operation is done 10 times for each topological map, to capture different combinations of nodes. During testing, we construct a full SPN for each test instance according to the algorithm described in section \ref{section:buildtmplspn}.

Similar to the experiments for the place grid method, we also occlude parts of the topological map, and ask the model to infer their attributes. Because the templates only consider connectivity of places, and place category is the only semantic attribute used as input for the template SPN, we do not expect the current template-based method to be able to infer a large occluded region accurately\footnote{This is the problem for place classification.}, but it should exhibit reasonable inference behavior in smaller scopes of the map. Therefore, we only randomly occlude several nodes per test instance. Also, because the training data is fundamentally different from the testing data (template versus full graph), we simply trained on the templates obtained from all sequences, and randomly picked a sequence to use its full topological map for testing. 

% Since there are nodes that are covered by during each partition operation, we initialize another SPN with dense structure which simply takes all unused nodes as inputs with random weights.

%% Besides, since we are ignoring several place categories such as kitchen and meeting room, there is a significant amount of nodes that would be all labeled unknown (the default label). Therefore, we test both cases of ignoring and not ignoring the unknown class during construction of topological maps examples.

\subsection{Results}\label{section:results}

\subsubsection{Place Grid Method}

\begin{figure}[!htb]
    \centering
    \captionsetup{width=.6\linewidth}
    \includegraphics[scale=0.4]{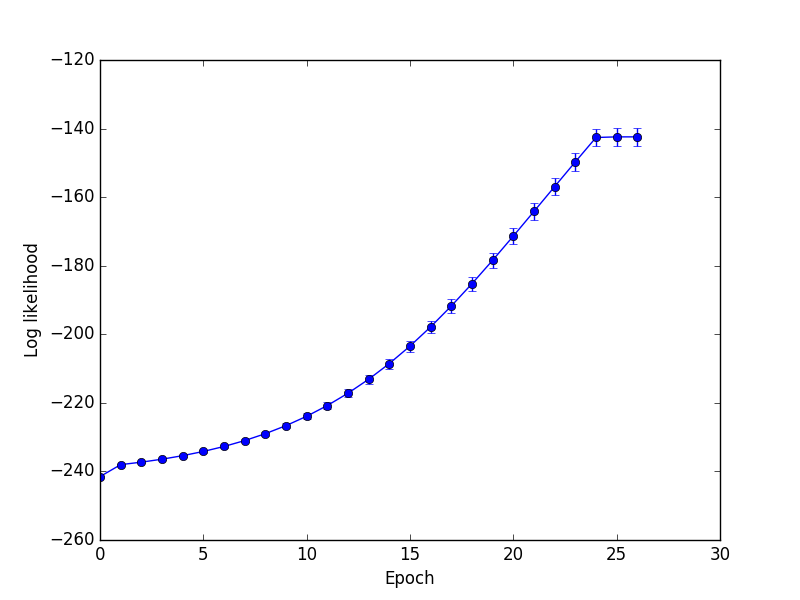}
    \caption{Average likelihood curve when training an SPN that models the placegrid of size 15 by 10. }
    \label{fig:likelihood}
\end{figure}

\begin{figure}[!htb]
    \centering
    \captionsetup{width=.8\linewidth}
    \includegraphics[scale=0.75]{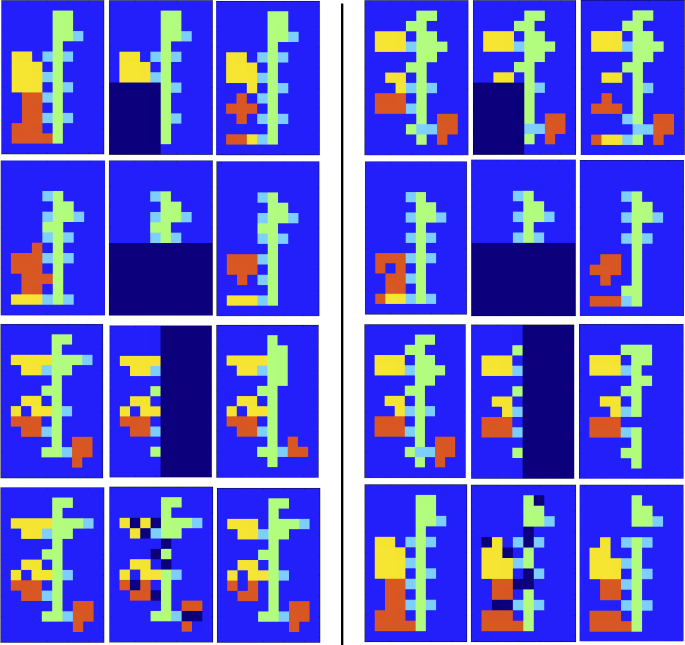}
    \caption{Results of map completion. On the left side of the division line are results from training samples, and on the other side are results from validation samples (i.e. on floors that the model were not trained on). For each triplet of grids, the left is groundtruth, the center is query, and the right is MPE inference result. The dark blue region or dots indicate places that are occluded. Color mapping: light blue = unknown, light green = corridor, cyan = doorway, yellow = small office, orange = large office.}
    \label{fig:basic_compl}
\end{figure}

\begin{figure}[!htb]
    \centering
    \captionsetup{width=.8\linewidth}
    \includegraphics[scale=0.7]{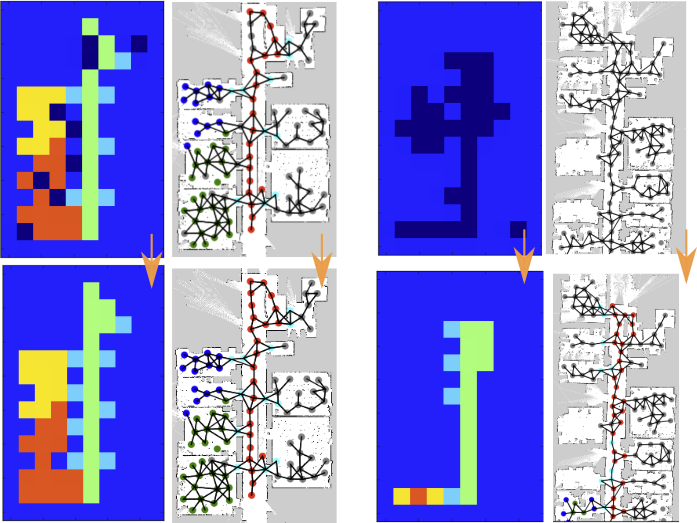}
    \caption{Showcase of the topological map gets updated as inference is done on the place grid. The top row shows query place grids (with dark blue as latent variables.), and the bottom row shows inference results produced by the network. In the topological map images, gray nodes have missing values. Color map in the topological map: red = corridor, cyan = doorway, blue = small office, green = large office.}
    \label{fig:pg_graph}
\end{figure}

\begin{figure}[!htb]
    \centering
    \captionsetup{width=.8\linewidth}
    \includegraphics[scale=0.4]{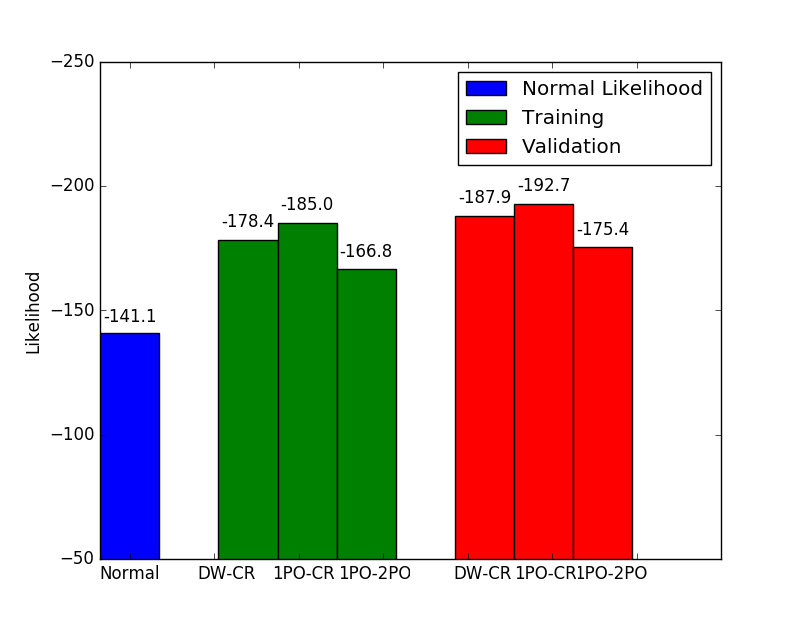}
    \caption{Novel structure detection results. On the x-axis, each label indicates the rooms that are swapped. For example, ``DW-CR'' means we swapped doorway and corridor classes. 1PO means small office. 2PO means large office. Note that the vertical axis is the log likehood with negative values.}
    \label{fig:novelty}
\end{figure}

\paragraph{Map completion} We tested our trained model's ability to infer missing information in the topological map. For each topological map test case, we occlude different parts of its projected place grid, and use the model to perform MPE inference to get an assignment for the missing values. The results are shown in figure \ref{fig:basic_compl}. The images on the left side are results from training samples, and those on the right side are results from validation samples. The first three rows show the results of map completion on place grid with different occluded regions. The last row show the results of map completion on place grid with randomly occluded cells. These results  indicate that the model learns the relations between place categories and general map structure to some degree. It is able to place doorway nodes between corridor and rooms. It is also able to learn that corridors extend along the same line, and different rooms categories do not mix up together. Errors in the inference do exist, as shown in several validation results where no doorway node is added between corridor and office.

We also demonstrate the actual use of the place grid method to fill in the spatial knowledge gaps in the topological graph in figure \ref{fig:pg_graph}. This figure shows two validation samples of the map completion task. On the left shows the completion of randomly occluded cells. On the right shows the completion of a place grid with an entirely occluded topological map. The graphs show the topological map with labels mapped from the corresponding place grids. The bottom row of images show the MPE inference results\footnote{For the case on the right, note that because we do not distinguish between certain place categories (such as kitchen) and the unknown, those skipped categories have the same label as the unknown. Therefore, the model believes that it is reasonable t  place a doorway cell between an unknown (background) cell and a corridor cell, because that background cell may refer to some category that we skipped.}. As we can see, the model can reasonably infer missing information, and it is possible to map from the inferred place grid to the original topological map to correct or fill in place categories.

\paragraph{Novel structure detection} One other way of checking the model's understanding of the environment is to see if it acts differently when two room classes switch their residing cells. Our result is shown in figure \ref{fig:novelty}. There are three groups of bars. The blue bar is the convergence likelihood during training (see figure \ref{fig:likelihood}). The green bars show the average likelihood using as input the place grids in training set where cells with certain place categories are swapped. The red bars show the same for place grids in validation set. From this plot, we can see that there is indeed a lower likelihood for unexpected swapping of place categories (e.g. between doorway and corridor). Also, it is reasonable that the likelihood after switching the two office classes (small and large) is higher than likelihoods of other, more irregular combinations of swapping. This shows that the model does learn the positional role that each room category plays. The slight decrease of likelihood in validation set is considered reasoanble. The bars in training (green) has similar pattern of  uneven length compared to those in validation (red), which indicates that the model exhibits consistent behavior for both training and validation data.

\subsubsection{Template-Based Method}

\begin{figure}[!htb]
    \centering
    \captionsetup{width=.8\linewidth}
    \includegraphics[scale=0.8]{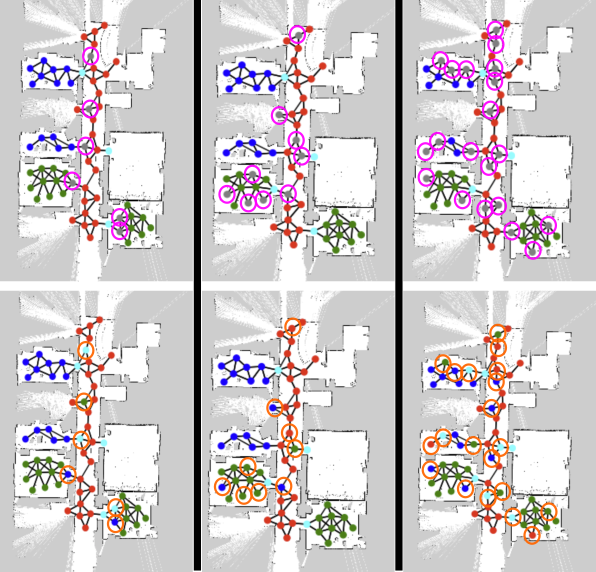}
    \caption{Results of template-based method on a sequence. Each column shows one test case of MPE query and its result. In each MPE query, topological map nodes are randomly occluded, indicated in gray nodes (magenta circles). The number of occluded nodes increase from left to right. Orange circles indicate the model's inference results of the occluded place categories.}
    \label{fig:tmpl_result}
\end{figure}

We conducted the missing value completion experiment on three sequences\footnote{The amount tested is limited by time, mostly due to the current expensive weight-sharing implementation.}. Figure \ref{fig:tmpl_result} shows the result of one of the sequences, where we skipped the places with unknown category, and we used a hierarchy of templates. We found that typically a template hierarchy leads to better results than using only a single template.
The constructed full SPN is able to infer place classes according to the surrounding information, but the performance is not very stable. It is able to infer that there is a doorway structure between two places of different classes, and that rooms have cluttered nodes of the same category. The unstable performance behaviors are mostly reflected in, for example, inferring a place to have class ``small office'' when its surroundings are mostly ``large office''. Figure \ref{fig:tmpl_result} also demonstrates this issue. 
%%%%%%%%%%%%%%%%%%%%%%%%%%%%%%
% Conclusion & Future work
%%%%%%%%%%%%%%%%%%%%%%%%%%%%%%
\newpage
\section{Conclusion \& Future Work}\label{section:conclusion}

In this thesis, we motivated the importance of learning a spatial representation for autonomous mobile robotics. We described the role of topological map and discussed its generation in the context of Deep Affordance Spatial Hierarchy. Then, we presented two methods that can enable learning of topological map using Sum-Product Networks. Finally, we described experiments and presented results that demonstrate the effect of the learning for each method.

Indeed, there is work to be done to improve each of the two methods. Especially for template-based method, the current method only consider connectivity of places and place categories as the only semantic attribute. We are looking to introduce more complexity in the templates and consider more information (such as laser range readings) when learning the template SPNs.

%%%%%%%%%%%%%%%%%%%%%%%%%%%%%%
% Acknowledgement
%%%%%%%%%%%%%%%%%%%%%%%%%%%%%%
\section{Acknowledgements}\label{section:acknowledgement}

I am sincerely grateful for the advice and encouragement from my advisor Dr.~Andrzej Pronobis, and insightful discussions I had we him as well as with my lab partner Kousuke Ariga. I also thank Yu Xiang, who shared plenty of his valuable experience in doing reserach projects. I thank the Robotics State-Estimation Lab run by Prof.~Dieter Fox for providing the space for me to do this work, and I enjoyed the atmosphere as well as the kind labmates there.

%%%%%%%%%%%%%%%%%%%%%%%%%%%%%%
% References
%%%%%%%%%%%%%%%%%%%%%%%%%%%%%%
\nocite{*}
\bibliographystyle{abbrv}
\bibliography{mythesis}

\newpage
\appendix

\end{document}